\documentclass[11pt]{article}

\usepackage[preprint]{acl}

\usepackage{times}
\usepackage{latexsym}

\usepackage[T1]{fontenc}
\usepackage[utf8]{inputenc}
\usepackage{microtype}
\usepackage{inconsolata}

\usepackage{graphicx}
\usepackage{adjustbox}

\usepackage{amsmath}
\usepackage{siunitx}
\usepackage{amssymb}
\usepackage{booktabs}
\usepackage{algorithm}
\usepackage{algorithmic}

\usepackage[dvipsnames]{xcolor}

\definecolor{groundedgreen}{RGB}{0,176,80}

\newcommand{\ourmethod}[1]{UniProbe}

\title{UniProbe: A Learnable Token-Level Hallucination Detector for Large VLMs using Multi-Structural Internal Representations}

\author{
\textbf{Dvir Samuel\textsuperscript{1} \quad
Guy Bar-Shalom\textsuperscript{2} \quad
Fabrizio Frasca\textsuperscript{2} \quad
Ethan Fetaya\textsuperscript{1,3}} \\
\textbf{Yftah Ziser\textsuperscript{1,4} \quad
Gal Chechik\textsuperscript{1,3} \quad
Haggai Maron\textsuperscript{1,2}} \\
\textnormal{\textsuperscript{1}NVIDIA Research, Tel Aviv, Israel} \\
\textnormal{\textsuperscript{2}Technion, Haifa, Israel} \\
\textnormal{\textsuperscript{3}Bar-Ilan University, Ramat Gan, Israel} \\
\textnormal{\textsuperscript{4}University of Groningen, Groningen, The Netherlands}
}

\begin{document}
\maketitle

\begin{abstract}
Large Vision-Language Models (LVLMs) achieve impressive visual reasoning and dialogue capabilities, yet frequently hallucinate content unsupported by the visual input. Effective mitigation requires token-level localization, enabling targeted intervention without discarding the entire response. Existing detectors require expensive full-model fine-tuning, rely on external verifiers that ignore the model's generation process, or reduce internal signals to isolated features and hand-crafted statistics, discarding spatial, sequential, and relational structure.
We introduce \textbf{\ourmethod{}}, a lightweight, unified, learnable detector that models a frozen LVLM's heterogeneous computational trace from a single forward pass. \ourmethod{} constructs a directed graph over image patches, query tokens, and generated tokens, with attention weights encoding their relations. It processes this trace with alternating structure-aware modules: a GNN for relational evidence, a ViT for 2-D visual geometry, and a GRU for response order. Interleaving them allows spatial, relational, and sequential evidence to interact throughout the detector.
We further develop a streaming variant for hallucination-aware decoding, which detects and resamples hallucinated tokens during generation, and a self-adaptation strategy aligning the detector with the LVLM's own generations. Across diverse LVLM backbones, \ourmethod{} achieves state-of-the-art token-level and object-hallucination detection. During decoding, it reduces object hallucinations by up to 55\% at $1.06\times$ the latency of standard generation. \href{https://research.nvidia.com/labs/par/uniprobe/}{Project Page}

\end{abstract}

\begin{figure}[t!]
\centering
\includegraphics[width=\columnwidth]{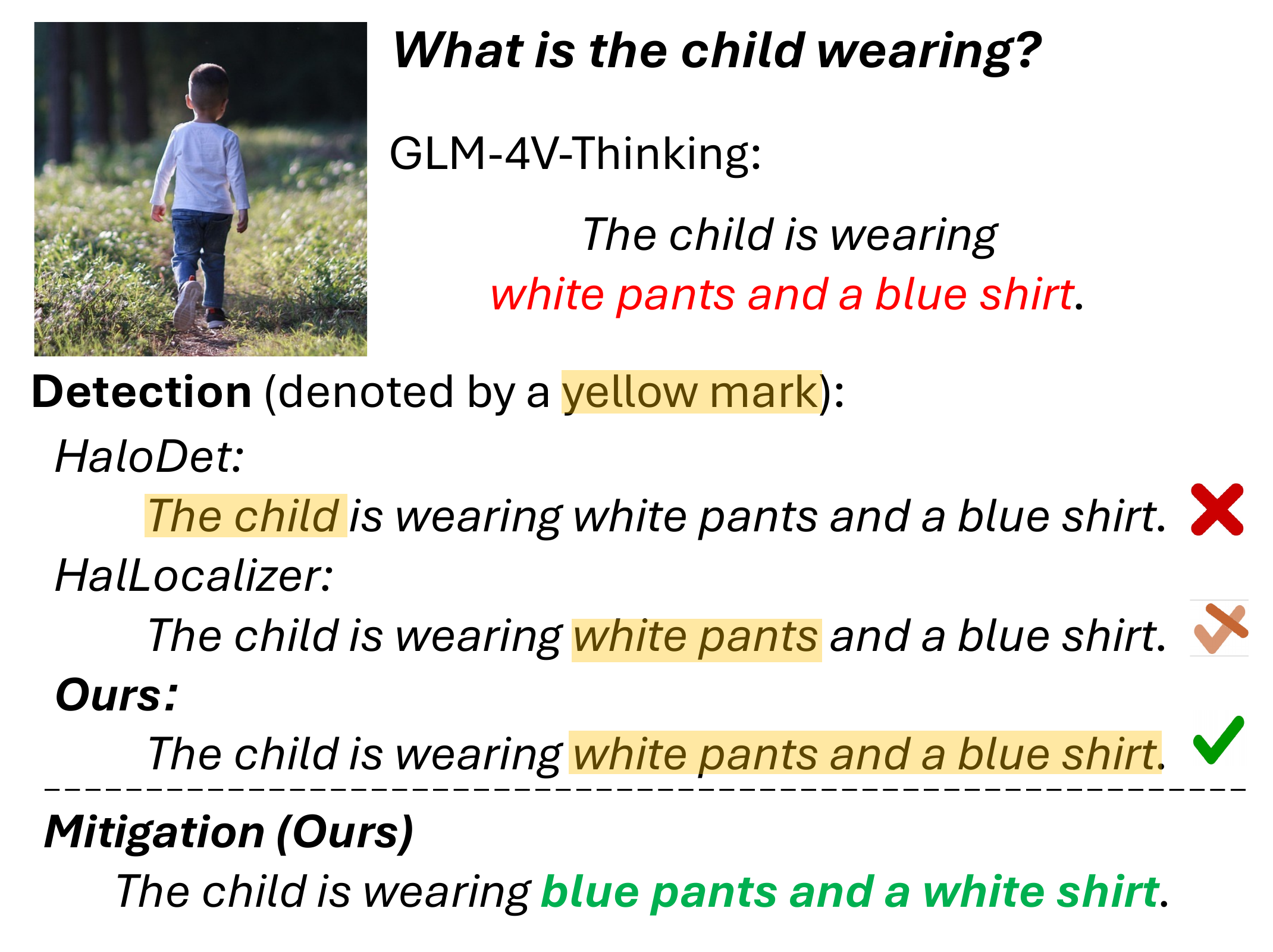}
  \caption{\textbf{Token-level hallucination detection and mitigation.}
  Given an input image, a LVLM generates hallucinated response texts (\textcolor{red}{red}).
  Existing SOTA detectors either misidentify non-hallucinated tokens (HaloDet) or fail to capture the full hallucination (HalLocalizer). Yellow highlights denote the tokens identified as hallucinated by each detector.
  Our method accurately \textit{detects} the hallucinated tokens and actively \textit{mitigates} them during generation to produce a grounded response (\textcolor{groundedgreen}{green}).}
\label{fig:fig1}
\end{figure}

\section{Introduction}

Large Vision-Language Models (LVLMs) \cite{llava, instructblip, internvl, glm4v} have achieved remarkable progress in bridging vision and language, enabling advanced visual reasoning, detailed image captioning, and interactive dialogue.
Despite their impressive capabilities, LVLMs remain notoriously prone to \emph{hallucinations}: generating text responses that are ungrounded in, or directly contradict, the visual input.
These unfaithful generations severely impair trustworthiness and hinder real-world applications.  Crucially, effective mitigation requires not only determining whether a response contains a hallucination, but also precisely identifying the affected tokens, enabling targeted intervention without discarding or regenerating the entire response. Accurate \textit{token-level hallucination detection} is therefore a central challenge in trustworthy multimodal AI.

Prior approaches to token-level hallucination detection generally fall into three categories: fine-tuning the LVLM to tag hallucinated spans \cite{mhalo,whitehead2}; training an external verifier that compares the response with the image \cite{halloc}; or probing the model's internal signals for evidence of grounding \cite{halp,tokengrounding,metatoken,dhcp}. Full fine-tuning updates billions of parameters and may degrade the model it is meant to safeguard, while external verification typically requires an additional vision-language model and ignores how the response was produced. Internal probes avoid both costs by reusing signals already computed during generation, but existing methods reduce these signals to isolated features or hand-crafted statistics, discarding much of their underlying structure.

Recent text-only (LLM) detectors \cite{losnet,actvit,charm} show that treating a model's computational trace as structured data, such as a sequence, tensor, or attention graph, is substantially more effective than approaches that use internal features but discard this structure. Extending this idea to LVLMs is challenging because their traces are inherently heterogeneous, combining a 2-D grid of image patches, 1-D query and response sequences, and cross-modal relations induced by attention.

We therefore introduce \textbf{\ourmethod{}}, a lightweight detector that reads this heterogeneous trace from a single forward pass of a frozen LVLM. It constructs a directed graph over image, query, and response tokens and processes it with alternating structure-aware modules: a GNN~\cite{gnn} for relational evidence, a ViT~\cite{dosovitskiy2021vit} for visual geometry, and a GRU~\cite{cho2014gru} for response order. Interleaving these modules allows spatial, relational, and sequential evidence to interact throughout the network while keeping the backbone LVLM frozen.

Beyond identifying hallucinations in completed responses, we use the same computational-trace formulation to intervene during generation. We train an online variant of \ourmethod{} that enables \emph{hallucination-aware decoding}: During decoding, each candidate token is assessed against the current computational trace. If the detector flags a token as hallucinated, we reject that token and resample an alternative, so the hallucination is corrected during generation and never reaches the final response.

Additionally, we identify and address an important practical challenge: hallucination detectors are trained on benchmarks that aggregate hallucinated responses from a variety of LVLM backbones, yet in real applications a detector must monitor and score its own generations. Because different backbones hallucinate in systematically different ways, a detector trained on other models' outputs faces a severe train–test distribution shift at deployment.
We mitigate this mismatch by adapting \ourmethod{} using responses sampled directly from the target model, aligning the detector with the host LVLM's specific generation dynamics.

We evaluate \ourmethod{} across a variety of open-source LVLM backbones on three tasks: token-level hallucination detection, object-hallucination detection, and real-time detection and mitigation of hallucinations in generated responses.
 It consistently outperforms state-of-the-art detectors on all three setups: token-level F1 improves by 4--6 points over the strongest trained baseline while keeping the backbone frozen, and object-hallucination detection on POPE rises from 41.0 to 63.1 F1. Finally, once adapted to the model's own outputs, our hallucination-aware decoding cuts object hallucinations by up to 55\% at just $1.06\times$ vanilla latency, without degrading response quality or substantially altering the model's output distribution.

\section{Related Work}

\subsection{Hallucination Detection from LLM Internals}
Recent work on LLMs has explored detecting hallucinations directly from their internal representations.
\citet{azaria2023internal} probe hidden states to predict statement truthfulness, and \citet{iti} steer
activations toward truthful behavior at inference time. Beyond single-vector probes, recent methods treat
the computational traces as structured data: attention maps reveal contextual hallucination
\cite{lookback,llmcheck}, ACT-ViT reads activation tensors with a vision transformer \cite{actvit}, CHARM
runs message passing on attention graphs \cite{charm}, and LOS-Net learns detectors over
output-distribution sequences \cite{losnet}.
The usefulness of internal representations extends beyond hallucination detection. Features extracted from
frozen diffusion models support personalized segmentation and retrieval \cite{pdm}, intermediate ViT
layers enable state-of-the-art place recognition \cite{effovpr}, and intermediate MLLM layers encode
substantial task-relevant information for video--text retrieval \cite{vidvec}. Together, these results
suggest that intermediate computations provide rich signals that can be exploited without modifying the
underlying model. We build on this perspective for multimodal hallucination detection, where
computational trace has a distinctive heterogeneous structure spanning image patches, query tokens, and
generated response tokens.

\subsection{Hallucination Detection in LVLMs}
LVLMs often generate fluent content that is unsupported by their visual input.
Existing approaches rely on external verification, full-model fine-tuning, or internal probing. \textit{HalLocalizer}~\cite{halloc} trains an external model to compare the image with the generated response, whereas \textit{HALP}~\cite{halp} probes the backbone's internal representations but predicts only a global hallucination label. \textit{HaloDet}~\cite{mhalo} and \citet{whitehead2} obtain token-level predictions by fine-tuning the LVLM, the latter optionally using pretraining on synthetically corrupted grounding data. In contrast, a growing family of methods flags object hallucinations from hand-crafted internal signals: the balance of attention between image and text tokens and its alignment with token representations \cite{tokengrounding,svar,pasCVPR26}, per-token confidence together with cross-modal attention patterns \cite{metatoken,dhcp}, or projections of visual features into the vocabulary space \cite{projectaway}.
A recent method, \textit{ZINA}~\cite{zina}, combines a hallucination detector with an external LVLM reviewer. Although effective, it requires a human-written reference caption at inference time, which is rarely available in real-world settings. In contrast, our method is fully reference-free.

Unlike all existing approaches, our lightweight learnable detector keeps the backbone frozen and treats its computational trace as a structured representation for general token-level hallucination localization.

\subsection{Hallucination Mitigation in LLMs}
A complementary line of work intervenes during or after decoding to mitigate hallucinations. OPERA~\cite{opera} penalizes over-trust attention patterns and backtracks when they occur; VCD~\cite{vcd} contrasts predictions from the original and distorted images to suppress language priors; and Woodpecker~\cite{woodpecker} corrects completed responses using external verifiers. These methods rely on fixed heuristics or additional models. Learned alternatives include Lookback Lens~\cite{lookback}, which guides text-only decoding with an attention-based detector, and TruthPrInt~\cite{truthprint}, which reads an LVLM's latent states but steers generation along a learned truthful direction. In contrast, our guardrail extends detector-guided decoding to the multimodal setting using a learned streaming detector over the full internal trace—not attention alone—and rejects and resamples the tokens it flags. This targeted intervention reduces hallucinations while preserving the model's output distribution.

\section{Notation and Problem Setup} \label{sec:setup}

We consider a frozen pre-trained large vision-language model $M$ (the  \emph{backbone}) that receives an image $I$ and a text query $x$, and autoregressively generates a response $r=(r_1,\dots,r_T)$ consisting of $T$ tokens. Internally, $M$ processes both inputs and outputs as a single sequence of $n$ tokens, segmented into 3 disjoint groups: \emph{image} patches, \emph{query} tokens, and \emph{response} tokens we wish to verify.

For a given generation, one can extract two primary signals from an intermediate layer $\ell$ of $M$: \textit{(1) Hidden States:} Let $h_i \in \mathbb{R}^{d}$ be the activation in the residual stream at position $i$, summarizing the model's computation for that token. \textit{(2) Attention Maps:} Let $A \in [0,1]^{n \times n}$ be the attention matrix (averaged across all heads), where $A_{ij}$ represents the attention weight from position $i$ to position $j$.

\textbf{Problem Statement.}
A token-level hallucination detection is defined as learning a detector $g_\theta$ that outputs a sequence of probabilities $p \in [0,1]^{T}$. Each $p_i$ represents the likelihood that token $r_i$ is a hallucination (i.e., not supported by $I$ or $x$). The detector is trained on responses annotated with hallucinated spans.

\section{Method}
\label{sec:method}

\begin{figure}[t!]
\centering
\includegraphics[width=\columnwidth]{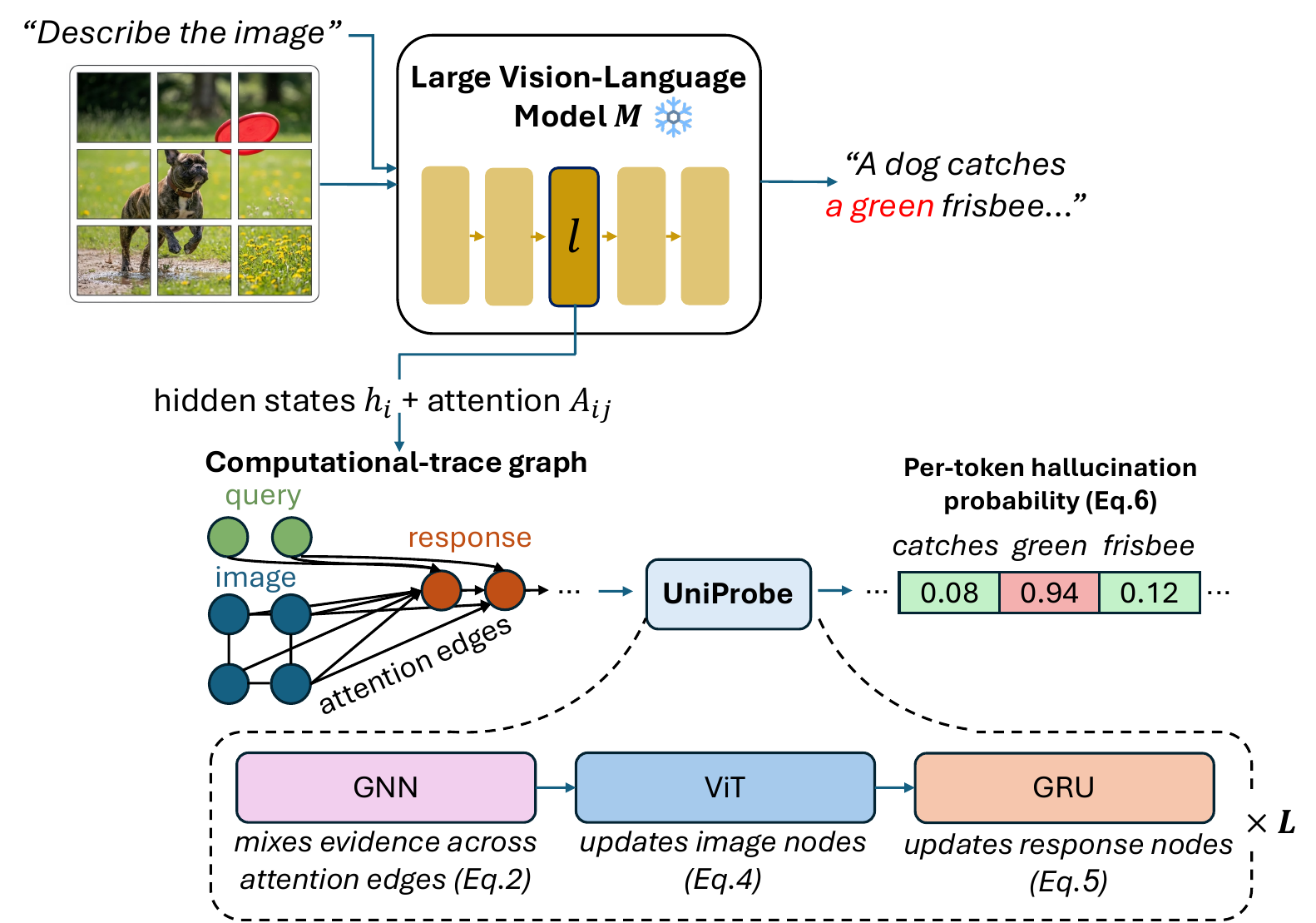}
\caption{\textbf{UniProbe architecture.} From a single forward pass of the frozen LVLM $M$, UniProbe reads layer-$\ell$ hidden states and attention to construct a \textit{computational-trace graph} over image, query, and response tokens. $L$ alternating blocks of GNN, ViT, and GRU, integrate cross-modal, spatial, and sequential evidence, followed by a linear head that predicts a hallucination probability for each response token.}

\label{fig:method}
\end{figure}

\textbf{Overview.} A single generation of $M$ yields three structured sources of information: the input \textbf{image}, represented as a 2-D grid of patches; the generated \textbf{response}, represented as a 1-D token sequence; and the \textbf{attention}, which captures interactions among image, query, and response tokens. Because these sources have different structures, we process each with a dedicated structure-aware module and couple them in a unified learnable detector built from \emph{alternating} blocks. Each block applies the three modules in turn to a shared per-token representation, allowing image, attention, and response evidence to interact throughout the network. The detector is trained jointly, end-to-end, while $M$ remains frozen. See Figure~\ref{fig:method} for an illustration.

\subsection{Computational-trace Graph} From one forward pass we build a directed, attributed graph $G=(V,E)$ that records, from $M$'s own internals, the visual and contextual evidence behind each emitted response token. The graph carries two complementary kinds of information. The first is \emph{internal signals} directly from $M$: the layer-$\ell$ hidden states and the attention weights. The second is lightweight \emph{structural metadata}, namely each node's type and, for image nodes, its patch coordinates; this metadata acts as a positional encoding providing each structure-aware module with the node’s position within the image or sequence.

\textit{Nodes} are tokens of three types: (1) all $T$ response tokens and a subset of (2) image patches and (3) query tokens.
Specifically, following the observation ~\cite{charm} that attention magnitude indicates which context a token relies on, we keep the graph lightweight by scoring each image position by the total attention it receives from the response, $s_j=\sum_{i\in\mathrm{resp}} A_{ij}$, and retaining the $N_{\mathrm{img}}$ highest-scoring positions. The same selection procedure is applied to query tokens. Node $v$ carries its layer-$\ell$ hidden state $h_v$ as a feature; image nodes
additionally carry a 2-D encoding $\gamma_v$ of their patch coordinates.

\textit{Edges} are derived directly from attention. For each response token $i$, we retain the strongest incoming connections from image patches, query tokens, and earlier response tokens $j<i$, with edge weight $a_{ij}=A_{ij}$. Specifically, we keep only the highest-$A_{ij}$ connections per token and drop the weak ones, which bounds the edge count. By causality, response-to-response edges originate only from preceding tokens, so the neighborhood of $i$ represents the visual and textual context used to generate $r_i$. We denote its type-$t$ neighbors by $\mathcal{N}_t(i)$, where $t\in\{\mathrm{img},\mathrm{query},\mathrm{response}\}$.

\subsection{The \ourmethod{} Architecture}
Given the computational-trace graph above, \ourmethod{} learns to combine its cross-modal, spatial, and sequential structure to identify hallucinated response tokens. We first project all node features into a shared representation space using type-specific linear maps:
\begin{equation}
x_i^{(0)} = W_{t(i)}\,h_i + \mathbf{1}[t(i){=}\mathrm{img}]\,\gamma_i ,
\end{equation}
where $t(i)$ is the type of node $i$. The network then stacks $L$ identical blocks; within block $b$ we
abbreviate $x_i\!\equiv\!x_i^{(b-1)}$. Each block runs the three modules in turn. \emph{First}, a
GNN~\cite{gnn} mixes evidence across modalities, updating every response node from its typed attention neighbors,
\begin{equation}
\begin{aligned}
m_i &= \sum_{t}\sum_{j\in\mathcal{N}_t(i)} a_{ij}\,W_t x_j + W_m c_i,\\[2pt]
\hat{x}_i &= \mathrm{LN}\big(x_i + \mathrm{ReLU}(W_s x_i + m_i)\big),
\end{aligned}
\label{eq:gnn}
\end{equation}
where every type $t$ has its own projection $W_t$, $W_s$ is a self-transform,
and $c_i\in\mathbb{R}^{3}$ stacks the incoming attention mass per modality,
\begin{equation}
c_i = \Big(\textstyle\sum_{j\in\mathcal{N}_t(i)} a_{ij}\Big)_{t\in\{\mathrm{img},\mathrm{qry},\mathrm{resp}\}},
\end{equation} a coarse summary of how much the token leaned on vision
vs.\ text. Separate $W_t$, for each modality, let the detector read ``attends to an image patch'' and ``attends to the query''
as distinct evidence. \emph{Then}, two within-modality modules capture spatial and sequential structure not explicitly modeled by the graph step: a ViT~\cite{dosovitskiy2021vit} over the image grid and a BiGRU~\cite{cho2014gru} along the response in generation order, both as residual
updates,
\begin{equation}
x_I \leftarrow \mathrm{LN}\big(\hat{x}_I + \mathrm{ViT}(\hat{x}_I)\big),
\end{equation}
\begin{equation}
x_R \leftarrow \mathrm{LN}\big(\hat{x}_R + \mathrm{GRU}(\hat{x}_R)\big),
\end{equation}

where $x_R\in\mathbb{R}^{T\times d}$ and $x_I\in\mathbb{R}^{N_{\mathrm{img}}\times d}$ stack the response- and image-node embeddings (where $d$ is the shared space dimension) and become
$x^{(b)}$.

Alternating the three modules $L$ times couples them into a single detector trained
jointly, end-to-end. A linear head on the final response embeddings gives the per-token scores
\begin{equation}
p_i = \sigma\!\big(w^{\top} x_i^{(L)}\big),\qquad i\in\{1,\dots,T\}.
\end{equation}
With the graph size bounded by fixed node and edge budgets, \ourmethod{} remains lightweight in practice and adds only modest overhead beyond the backbone computation it reads from.

\subsection{Online Detection and Hallucination-aware Decoding}
\label{sec:streaming}
We turn \ourmethod{} into a streaming (online) detector by simply replacing its \textit{bidirectional response GRU} with a \textit{unidirectional one}. Because the graph already follows generation order and the ViT processes only the static image, this change ensures that the score of $r_i$ depends only on the prefix $r_{\le i}$, without otherwise modifying the architecture.

The streaming detector further turns detection into \emph{prevention}. As $M$ decodes, we score each newly generated token using the causal computational trace. If its hallucination probability exceeds a threshold $\tau$, we reject it, banning its first token and re-decoding, following the detector-guided
decoding of \citet{lookback}. Otherwise we accept and continue.

\subsection{Mitigating Distribution Shift in Self-generation}

Current token-level datasets  \cite{mhalo,halloc} share a fixed protocol: the detector is supervised on hallucination annotations of responses produced by a set of \emph{other} LVLMs, and is then run over $M$ while $M$ is \emph{teacher-forced} on those same pre-collected responses. In real deployment, however, the detector must judge $M$'s \emph{own}, freely sampled generations, which are more fluent and confident than the third-party responses seen in training. This train--test distribution shift lowers detection precision dramatically.

We address this gap through self-adaptation, without requiring manual hallucination annotations. We sample a small subset of Objects365~\cite{object365} images with ground-truth object annotations, generate free-form captions using the target model $M$, and automatically label each generated object mention using CHAIR~\cite{chair}. Specifically, an object mention is marked as hallucinated if it does not correspond to any ground-truth object in the image. We then further fine-tune \ourmethod{} on these automatically labeled generations. This aligns our detector with $M$'s own generations, thereby reducing the distribution shift.

\section{Experiments}
\label{sec:exp}

We evaluate our approach on three setups: (1) \textbf{token-level hallucination detection}, on both teacher-forced and self-generated responses; (2) \textbf{object hallucination detection}; and (3) \textbf{hallucination-aware decoding during streaming generation}.

\textbf{Datasets and metrics.}
\emph{(1) Token-level:} MHALO~\cite{mhalo}, a fine-grained detection benchmark whose responses are annotated with hallucinated spans (five eval sets, $\sim$500 samples each, spanning general, knowledge, and reasoning prompts); and HalLoc~\cite{halloc}, a large-scale localization benchmark with per-token labels in several categories (object, attribute, relationship, scene) across VQA, instruction, and captioning.

We report F1\textsubscript{M}, a word-overlap F1 between predicted and GT hallucination spans; F1\textsubscript{IoU}, the mean over gold spans of the best word-level IoU with any predicted span; and IF, the fraction of well-formed predictions. On HalLoc we report per-category Precision/Recall/F1 following \cite{halloc}.

\emph{(2) Object hallucination detection:} We evaluate object presence on POPE~\cite{pope}, following the same training data and evaluation protocol as \citet{tokengrounding,projectaway}. Specifically, we ask the detector to identify incorrect answers rather than directly scoring the LVLM's responses, and report F1 and AUC.

\emph{(3) Self-generated captions:} We evaluate detection on the model's own COCO~\cite{lin2014coco} captions using CHAIR~\cite{chair}. CHAIR\textsubscript{i} measures the fraction of mentioned objects absent from the image, and CHAIR\textsubscript{s} the fraction of captions containing at least one such object. Labels are generated automatically by matching caption words to COCO categories and synonyms and comparing them with GT annotations, without a judge model. We also report detection F1 against these labels.

\textbf{Backbones.} We test our detector across various LVLMs: GLM-4.1V-9B~\cite{glm4v}, Qwen-3-VL~\cite{qwen3}, LLaVA-1.5-7B~\cite{llava}, InternVL2-8B~\cite{internvl}, and
InstructBLIP-7B~\cite{instructblip}.

\begin{table}[t]
\centering
\begin{adjustbox}{max width=\columnwidth}
\setlength{\tabcolsep}{4pt}\small
\begin{tabular}{@{}lccc@{}}
\toprule
\textbf{MHALO} & F1\textsubscript{M} & F1\textsubscript{IoU} & IF \\
\midrule
\multicolumn{4}{l}{\textit{Zero-shot prompted LVLMs}}\\
MiniCPM-V 2.6              & 18.4 & 13.1 & 82.1 \\
InternVL2-76B             & 28.5 & 21.2 & 88.1 \\
Qwen-VL-Max               & 30.4 & 22.9 & 99.7 \\
Llama-3.2-90B-V           & 34.9 & 23.6 & 94.1 \\
Abab7-chat                & 36.9 & 28.0 & 95.2 \\
Claude-3.5-Sonnet   & 49.8 & 30.1 & 98.9 \\
GLM-4V-Plus               & 35.9 & 30.3 & 94.2 \\
Gemini-1.5-Pro     & 51.9 & 36.7 & 99.0 \\
GPT-4o              & 56.8 & 40.6 & 99.2 \\
Claude-4.8-Opus              & 58.6 & 43.9 & 99.3 \\
\midrule
\multicolumn{4}{l}{\textit{Trained detectors (Qwen-3-VL backbone)}}\\
Whitehead et al.~\cite{whitehead2} & 49.7 & 45.1 & 99 \\
HalLocalizer~\cite{halloc} & 49.9 & 45.3 & \textbf{100} \\

HaloDet~\cite{mhalo}    & 55.4 & 46.7 & 91 \\
\textbf{\ourmethod{} (Ours)}             & \textbf{61.7} & \textbf{51.2} & \textbf{100} \\
\midrule
\multicolumn{4}{l}{\textit{Trained detectors (GLM-4V backbone)}}\\
Whitehead et al.~\cite{whitehead2} & 56.0 & 47.3 & 99 \\
HalLocalizer~\cite{halloc} & 56.0 & 47.9 & \textbf{100} \\

HaloDet~\cite{mhalo}    & 59.1 & 49.8 & 92 \\
\textbf{\ourmethod{} (Ours)}             & \textbf{63.2} & \textbf{52.9} & \textbf{100} \\
\bottomrule
\end{tabular}
\end{adjustbox}
\caption{\textbf{Token-level detection on MHALO.} Our approach beats zero-shot frontier LVLMs and SoTA trained detectors on all metrics and across different backbones.}
\label{tab:mhalo}
\end{table}

\textbf{Baselines.} We compare against baselines for \textit{token-level detection}, \textit{object-level detection}, and \textit{hallucination-aware decoding}. For \emph{token-level detection}, we include zero-shot frontier LVLMs using Analyze-then-Judge~\cite{mhalo}, as well as trained methods. \emph{HaloDet}~\cite{mhalo} fine-tunes the backbone to regenerate responses with inline hallucination tags, while \citet{whitehead2} replaces the LM head with a token classifier. \emph{HalLocalizer}~\cite{halloc} instead trains an external verifier over the image and response.

For \emph{object-hallucination detection}, we also compare with methods that read signals from a frozen backbone. \emph{HALP}~\cite{halp} probes pre-generation representations at the sample level; token-wise probing of response hidden states is covered by our sequence-only ablation (Sec.~\ref{sec:abl}). \emph{MetaToken}~\cite{metatoken} and \emph{DHCP}~\cite{dhcp} train lightweight classifiers over confidence and cross-modal attention features. \emph{SVAR}~\cite{svar}, \emph{Token Grounding}~\cite{tokengrounding}, and \emph{PAS}~\cite{pasCVPR26} use hand-crafted attention or representation-alignment scores, while \emph{ProjectAway}~\cite{projectaway} projects visual features into the language space.

For \emph{hallucination-aware decoding}, we apply the same guardrail with HaloDet~\cite{mhalo} and PAS\cite{pasCVPR26}.

\begin{table}[t]
\centering
\begin{adjustbox}{max width=\columnwidth}
\setlength{\tabcolsep}{4pt}\small
\begin{tabular}{@{}llccc@{}}
\toprule
\textbf{HalLoc} &  & P & R & F1 \\
\midrule
\emph{any}   & Always-1      & 10.1 & 100  & 16.6 \\
\midrule
InternVL2    & HalLocalizer   & 69.5 & 73.1 & 70.6 \\
             & \textbf{\ourmethod{} (Ours)} & \textbf{73.8} & \textbf{75.2} & \textbf{73.9} \\
\midrule
LLaVA-1.5    & HalLocalizer   & 70.8 & 71.1 & 69.8 \\
             & \textbf{\ourmethod{} (Ours)} & \textbf{74.6} & \textbf{75.0} & \textbf{74.3} \\
\midrule
InstructBLIP & HalLocalizer   & 64.1 & 68.9 & 65.3 \\
             & \textbf{\ourmethod{} (Ours)} & \textbf{65.1} & \textbf{71.1} & \textbf{67.4} \\
\bottomrule
\end{tabular}
\end{adjustbox}
\caption{\textbf{Token-level localization on HalLoc.} Ours beats the trained HalLocalizer on all three
backbones (overall P/R/F1 over the 12 subset$\times$category cells).}
\label{tab:halloc}
\end{table}

\begin{table}[t]
\centering
\small
\begin{adjustbox}{max width=\columnwidth}
\setlength{\tabcolsep}{4pt}\small
\begin{tabular}{@{}lc|cc @{}}
\toprule
& \multicolumn{1}{c}{COCO} & \multicolumn{2}{c}{POPE} \\
\cmidrule(lr){2-2}\cmidrule(lr){3-4}
Method & F1 & F1 & AUC\\
\midrule
MetaToken-GB~\cite{metatoken}       & 68.0 & 22.0 & 73.0 \\
HalLoc~\cite{halloc}                & 71.0 & 29.0 & 71.0 \\
SVAR~\cite{svar}                    & 76.0 & 25.0 & -- \\
DHCP~\cite{dhcp}                    & 73.0 & 27.0 & 69.0 \\
ProjectAway~\cite{projectaway}      & 79.0 & 38.0 & 70.0\\
HALP (flat probe)~\cite{halp}                    &  -- & 39.5 & 68.9\\
Token Grounding~\cite{tokengrounding}
                                      & 82.0 & 41.0 & 75.0 \\

\textbf{\ourmethod{} (Ours)}                       & \textbf{92.3} & \textbf{63.1} & \textbf{90.0}\\
\bottomrule
\end{tabular}
\end{adjustbox}
\caption{\textbf{Self-generation and object-hallucination detection.} \ourmethod{} outperforms prior detectors on self-generated COCO captions and POPE across all metrics.}
\label{tab:object_detection}
\end{table}

\begin{table}[t]
\centering
\begin{adjustbox}{max width=\columnwidth}
\setlength{\tabcolsep}{4pt}\small
\begin{tabular}{@{}lcccccc@{}}
\toprule
 & \multicolumn{1}{c}{Detection} & \multicolumn{5}{c}{Mitigation} \\
\cmidrule(lr){2-2}\cmidrule(lr){3-7}
Method & $F1_{obj}$ & CHAIR\textsubscript{i} & CHAIR\textsubscript{s} & JS Div & SPICE & Lat. \\
\midrule
Vanilla                                 & ---           & 18.0          & 37.2          & --     & 0.214          & 1.00$\times$\\
VCD~\cite{vcd}                     & --          & 16.8          & 36.1          & 0.031  & 0.203          & 2.00$\times$\\
PAS~\cite{pasCVPR26}                     & 24.9          & 17.1          & 35.8          & 0.029  & 0.205          & 1.14$\times$\\
HaloDet~\cite{mhalo}                     & 11.2          & 17.9          & 37.0          & 0.0004 & 0.214          & 1.30$\times$\\
\quad{}+ FT on self-generation           & 34.1          & 13.1          & 27.4          & 0.011  & 0.213          & 1.30$\times$\\
\textbf{Ours (streaming)}                & \textbf{32.6} & \textbf{15.3} & \textbf{31.4} & 0.012  & 0.218          & \textbf{1.06$\times$}\\
\textbf{\quad{}+ FT on self-generation}  & \textbf{63.8} & \textbf{8.2}  & \textbf{16.6} & 0.013  & \textbf{0.222} & \textbf{1.06$\times$}\\
\bottomrule
\end{tabular}
\end{adjustbox}
\caption{\textbf{Detection and mitigation on streaming self-generated captions on COCO.} The same streaming
guardrail driven by different detectors: our approach reduces object hallucination the most, at $1.06\times$ vanilla latency, while preserving caption quality (SPICE) and the output distribution.}

\label{tab:decode}
\end{table}

\begin{figure*}[t!]
\centering
\includegraphics[width=\linewidth]{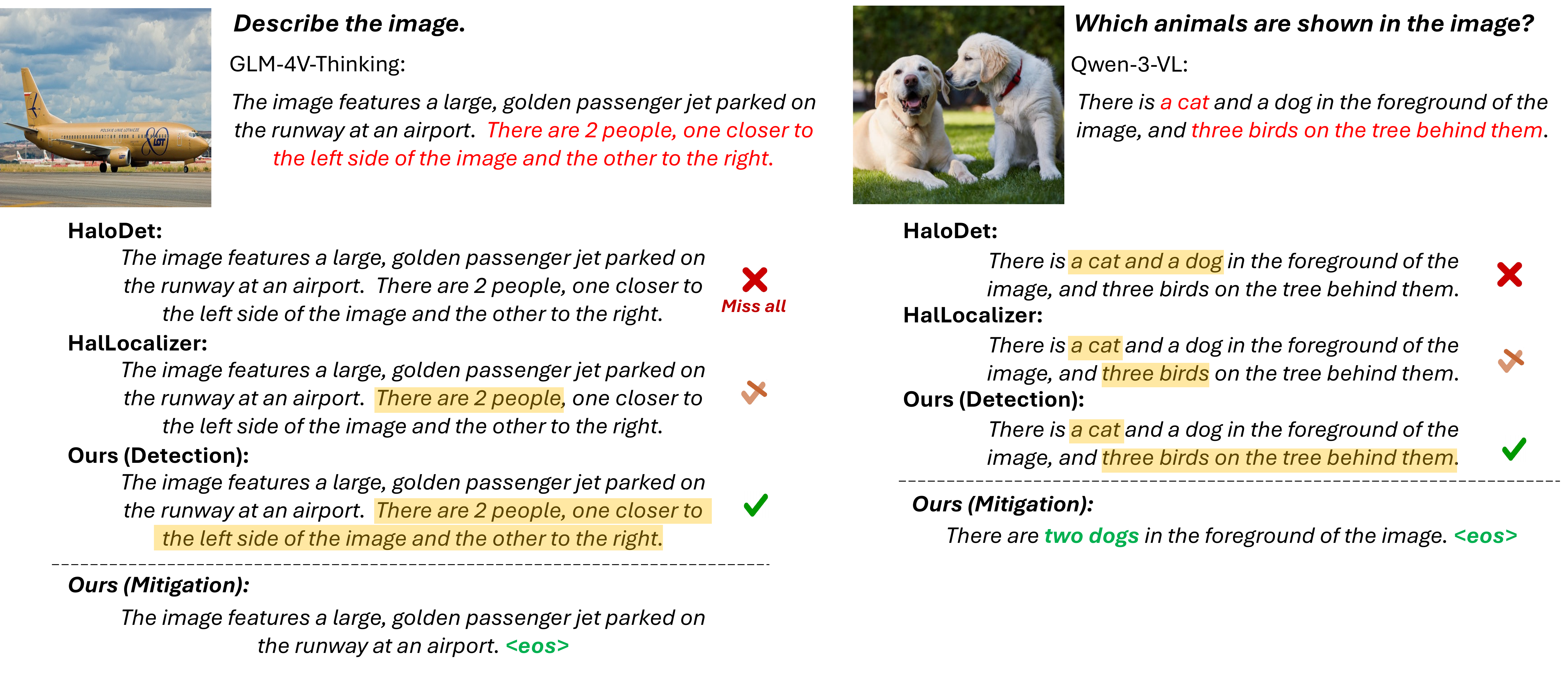}
\caption{\textbf{Qualitative token-level detection and mitigation across different LVLM backbones.}
Existing SOTA methods exhibit different failure modes: they incorrectly flag visually grounded text, detect only part of the hallucinated span, or miss it entirely. In contrast, \ourmethod{} accurately localizes the hallucinated tokens in both examples and uses its predictions during hallucination-aware decoding to produce the grounded outputs shown in \textcolor{groundedgreen}{green}. Each column presents an example from a different backbone, demonstrating that the observed behavior is not model-specific. Backbone-generated hallucination is shown in \textcolor{red}{red}, while yellow highlights mark the tokens predicted as hallucinated by each detector.}
\label{fig:qualitative}
\end{figure*}

\subsection{Results}

\textbf{Token-level detection.} Table~\ref{tab:mhalo} shows that zero-shot prompting of frontier LVLMs is
weak at fine-grained detection (F1\textsubscript{IoU} 40.6 for GPT-4o, 43.9 for Claude-4.8-Opus). Trained
detectors close part of the gap, but our frozen read-out beats them all on both backbones: F1\textsubscript{M}/F1\textsubscript{IoU}
63.2/52.9 vs.\ 59.1/49.8 for HaloDet on GLM-4V, and 61.7/51.2 vs.\ 55.4/46.7 on Qwen-3-VL. This holds even though
HaloDet fine-tunes every backbone weight and ours updates none. The advantage carries over to HalLoc
(Table~\ref{tab:halloc}), where \ourmethod{} beats the trained HalLocalizer on all three backbones.

The accuracy gains come with modest computational overhead. Relative to vanilla generation ($1.00\times$), \ourmethod{} scores the full response at $1.15\times$ latency, compared with $1.30\times$ for HaloDet, $1.21\times$ for HalLocalizer, and $1.23\times$ for Whitehead et al. Thus, \ourmethod{} provides the strongest accuracy--efficiency trade-off, achieving higher detection performance while adding less latency than all trained baselines.

Figure~\ref{fig:qualitative} qualitatively illustrates the advantage of \ourmethod{} on generations produced by different LVLM backbones. The examples expose several limitations of existing token-level detectors: they can over-detect by marking visually supported content, under-detect by identifying only a short portion of a hallucinated statement, or fail to detect the hallucination altogether. In contrast, \ourmethod{} consistently recovers the complete unsupported span across both backbones. Importantly, this improved localization translates directly into more effective intervention: when used for hallucination-aware decoding, the same detector prevents the unsupported continuation and yields a response that remains grounded in the visual input.

\textbf{Object hallucination detection.} Table~\ref{tab:object_detection} evaluates the detection of hallucinated object mentions with LLaVA-1.5-7B, both in self-generated COCO captions and among the model's POPE answers. \ourmethod{} reaches 92.3 F1 on COCO captions and 63.1 F1 / 90.0 AUC on POPE, compared to 82.0 and 41.0 / 75.0 for Token Grounding, the strongest prior detector. All prior detectors plateau at 69--75 AUC on POPE, whether they are external verifiers or flat, hand-crafted read-outs of the same frozen internals; structuring that same evidence as a computational trace lifts detection to 90.0 AUC on this heavily imbalanced task, where the wrong answers to be caught are rare.

\begin{table}[t]
\centering
\setlength{\tabcolsep}{1pt}\small
\begin{minipage}[t]{0.4\columnwidth}\centering
{(a) Each module alone}\\[2pt]
\begin{tabular}{@{}lcc@{}}
\toprule
 & F1\textsubscript{M} & F1\textsubscript{IoU}\\
 \midrule
 MLP (flat probe) & 31.9 & 25.7 \\
  Transformer (flat probe) & 33.2 & 27.8 \\
\midrule
ViT only      & 51.8 & 45.1\\
GNN only      & 56.9 & 49.8\\
GRU only      & 53.1 & 46.7\\
\textbf{Ours} & \textbf{63.2} & \textbf{52.9}\\
\bottomrule
\end{tabular}
\end{minipage}\hfill
\begin{minipage}[t]{0.4\columnwidth}\centering
{(b) Remove one module}\\[3pt]
\begin{tabular}{@{}lcc@{}}
\toprule
 & F1\textsubscript{M} & F1\textsubscript{IoU}\\
\midrule
\textbf{Ours} & \textbf{63.2} & \textbf{52.9}\\
$-$ ViT       & 61.5 & 52.5\\
$-$ query     & 60.2 & 51.7\\
$-$ BiGRU     & 58.6 & 51.6\\
$-$ graph     & 53.1 & 46.7\\
\bottomrule
\end{tabular}
\end{minipage}
\caption{\textbf{Architecture and component ablations on MHALO.} (a)  MLP and Transformer on flat probes and individual modules underperform the full model. (b) Removing any component hurts, with the graph and response modules contributing most.}
\label{tab:abl}
\end{table}

\textbf{Streaming detection and hallucination-aware decoding.}
We evaluate deployment on GLM-4.1V's freely generated captions, with hallucinated object mentions labeled by CHAIR. The streaming detector achieves $32.6$ $F1_{\mathrm{obj}}$ before adaptation and $63.8$ after self-adaptation, while retaining its MHALO performance at $63.5/53.0$ $F1_{\mathrm{M}}/F1_{\mathrm{IoU}}$ with IF $=100$.

Table~\ref{tab:decode} applies the same reject-and-resample guardrail to every detector, isolating the effect of detection quality. PAS and HaloDet, with $24.9$ and $11.2$ $F1_{\mathrm{obj}}$, only modestly reduce CHAIR\textsubscript{i}/CHAIR\textsubscript{s} to $17.1/35.8$ and $17.9/37.0$, while adding more latency. Our unadapted streaming detector lowers them from $18.0/37.2$ to $15.3/31.4$, and self-adaptation further reduces them to $8.2/16.6$, a $55\%$ reduction, at only $1.06\times$ vanilla latency. Adapting HaloDet raises its $F1_{\mathrm{obj}}$ to $34.1$ and lowers CHAIR to $13.1/27.4$, but it remains well behind ours, showing that adaptation alone does not explain the gain. The improvement also does not come from suppressing content: guarded captions retain $95.4\%$ of the vanilla word count ($143.7$ vs.\ $150.7$), preserve correct-object coverage ($2.14$ vs.\ $2.23$), and eliminate the six empty captions produced by vanilla decoding. Perplexity remains below vanilla's ($2.27$ vs.\ $2.30$), with only a small distribution shift ($\mathrm{JS}=0.013$). A complementary user study appears in the appendix.

\begin{figure}[t]
\centering
\includegraphics[width=0.7\columnwidth]{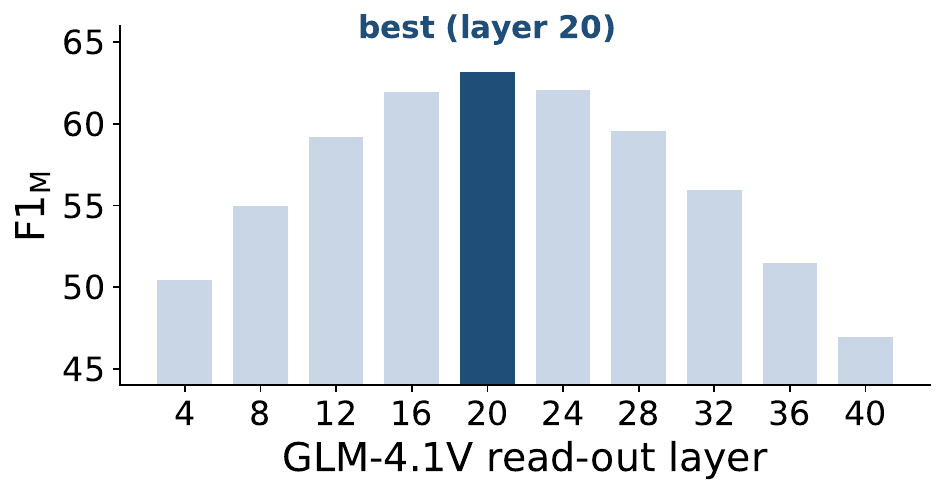}\\
\includegraphics[width=0.7\columnwidth]{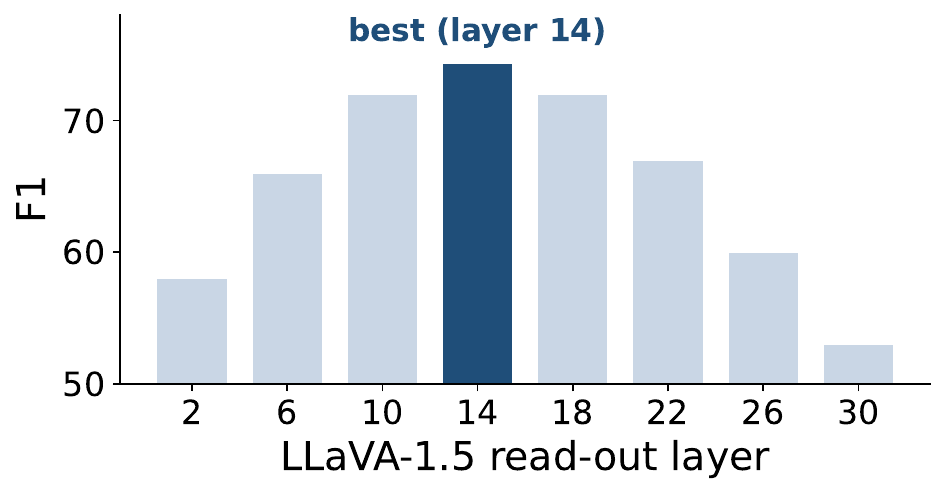}
\caption{\textbf{The grounding signal lives in the middle layers, across backbones.} Detection F1 vs.\ the
read-out layer: GLM-4.1V (top, MHALO) peaks at layer~$20$ and LLaVA-1.5 (bottom, HalLoc) at layer~$14$, both
degrading toward the output.}
\label{fig:layer}
\end{figure}

\section{Ablation Study}
\label{sec:abl}
We analyze the key design choices underlying \ourmethod{} through ablations of its architecture and hallucination-aware decoding. Unless stated otherwise, we evaluate the detector on MHALO with the GLM-4V backbone; additional ablations are provided in the appendix.

\textbf{Architecture and components.} Table~\ref{tab:abl} compares \ourmethod{} with flat probes and ablates its structure-aware modules. Flat MLP and Transformer probes perform substantially worse, trailing the full model by at least $30.0$ F1\textsubscript{M}, highlighting the importance of preserving the trace's heterogeneous structure. Among individual modules, the GNN performs best but reaches only $56.9$ F1\textsubscript{M}, compared with $63.2$ for the full model. Conversely, removing any component hurts, with the largest drops caused by removing the relational graph ($-10.1$) and response recurrence ($-4.6$). These results show that hallucination evidence is distributed across the relational, spatial, and sequential structure of the trace and is most effective when jointly modeled.

\textbf{Read-out layer.} \ourmethod{} reads from a single mid-network layer. Fig.~\ref{fig:layer} sweeps this choice on GLM-4.1V: detection climbs through the early layers, peaks at layer~$20$, and drops toward the output, where representations specialize for next-token prediction rather than grounding. The other backbones behave the same way, peaking mid-network (layer~$14$ of~$32$ for LLaVA-1.5 and InternVL2); the peak layer is chosen per backbone on validation data. Reading and training from several layers at once, consistently underperforms the single peak layer.
Because grounding is concentrated in a narrow mid-network band, depth pooling dilutes it with early generic features and late next-token-prediction features. Varying activation scales further bias naive averaging toward high-norm, less-informative layers.

\begin{figure}[t]
\centering
\includegraphics[width=0.7\columnwidth]{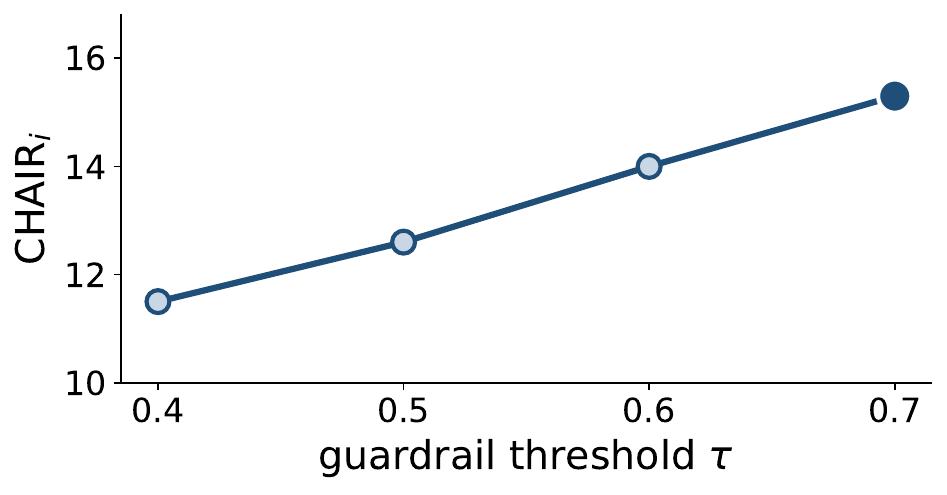}
\caption{\textbf{The threshold controls how much hallucination the guardrail removes.} CHAIR$_i$ on
GLM-4.1V falls as $\tau$ tightens; we operate at $\tau{=}0.70$ (marked), the most aggressive setting that
still leaves the correctly mentioned objects intact (lower $\tau$ removes more but starts deleting correct
content)
}
\label{fig:front}
\end{figure}

\textbf{Decoding threshold.} The threshold $\tau$ trades hallucination reduction against content preservation (Fig.~\ref{fig:front}). We use $\tau\!=\!0.70$, the most aggressive setting that reduces hallucinations without removing correctly generated content. Lowering it to $\tau\!=\!0.40$ reduces CHAIR\textsubscript{i} by $36\%$, but also removes $12\%$ of correct mentions.

\section{Conclusion}

We introduced \ourmethod{}, a lightweight learnable detector that models the internal attention, visual, and response representations of a frozen LVLM. Across multiple backbones and benchmarks, it improves token-level and object-level hallucination detection while keeping the underlying model frozen. We also propose a streaming variant that further reduces object hallucinations during decoding while preserving response quality and adding little latency. These results show that structured internal model traces provide a practical signal for both hallucination detection and mitigation.

\section*{Limitations}
Despite its effectiveness, our approach has several limitations. It requires access to hidden states and attention maps, and applicable only to open-source models or directly by closed-source providers. We also train a separate readout for each backbone, as internal representations differ across models, and leave cross-backbone transfer to future work.

\bibliography{custom}

\clearpage
\appendix

\section*{Appendix}

\section{Implementation Details}
 We extract internal signals from a single mid-network layer of each frozen backbone, chosen per backbone by validation F1 over a coarse layer sweep: layer 20 for GLM-4.1V and InstructBLIP-7B, and layer 14 for LLaVA-1.5 and InternVL2.
\ourmethod{} uses a hidden width of $h\!=\!256$ and $L\!=\!2$ alternating blocks across all backbones ($\sim$16M trainable parameters). Each block applies one message-passing GNN layer, a single-layer bidirectional GRU over the response tokens, and one ViT encoder layer over the image nodes ($4$ attention heads, feed-forward width $2h$); dropout is $0.2$ ($0.1$ within the ViT layers). Pruning is done following \cite{charm}.
We train for $8$ epochs with per-graph updates (one graph per step) using Adam with a learning rate of $10^{-3}$, weight decay of $10^{-4}$, and class-balanced binary cross-entropy.
For hallucination-aware decoding, we use the streaming detector with a threshold $\tau\!=\!0.7$. The threshold is applied to the detector probability $p_i$ and is selected on a held-out development set.
Unless otherwise noted, all architectural, graph-construction, training, and decoding hyperparameters reported below were selected through ablation studies on held-out validation data. For self-adaptation, we sample the target model's own captions on $500$ Objects365 images. All experiments, for both \ourmethod{} and the trainable detectors are conducted on a single H100 GPU.

\section{User Study}
Automatic object-hallucination rates (Table~\ref{tab:decode}) measure whether the guardrail removes ungrounded objects, but not whether the resulting caption reads, to a person, as a better description of the image. We therefore complement Table~\ref{tab:decode} with a human preference study over the mitigated captions of the three guardrail-driven methods: \ourmethod{}, PAS, and HaloDet.

\paragraph{Protocol.}
We assemble $100$ comparisons; each shows an image alongside the mitigated caption produced by each of the three methods, presented in random order and without method labels. $100$ raters recruited through Mechanical Turk each judge $20$ comparisons drawn at random from this set ($2{,}000$ judgments in total). Participants were required to be fluent in english and  have an approval rate above 90\%. Each participant received 10 cents for each rating.

\paragraph{Instructions.}
Participants received the following instruction: ``For each image, you will see three captions presented in random order. Select the caption that best describes the visible content of the image. Consider factual accuracy and completeness, and do not prefer a caption solely because it is longer or more fluent.'' Each comparison required exactly one selection.

\paragraph{Consent and ethics.}
Before participating, raters were informed that their anonymous judgments would be used for research and provided consent. No personally identifying information was collected or retained.

\paragraph{Results.}
Table~\ref{tab:userstudy} reports the outcome. Raters prefer \ourmethod{}'s mitigated caption in $55\%$ of judgments, against $30\%$ for PAS and $15\%$ for HaloDet. Human preference thus mirrors the automatic CHAIR ranking of Table~\ref{tab:decode}: the detector that localizes hallucinations most precisely also produces the captions people find most faithful, by a wide margin.

\begin{table}[t!]
\centering\small
\begin{tabular}{@{}lc@{}}
\toprule
Method & Preferred (\%) \\
\midrule
\textbf{\ourmethod{} (Ours)} & \textbf{55} \\
PAS~\cite{pasCVPR26} & 30 \\
HaloDet~\cite{mhalo} & 15 \\
\bottomrule
\end{tabular}
\caption{\textbf{Human preference study} on mitigated captions ($100$ raters $\times$ $20$ comparisons $=2{,}000$ judgments). Fraction of judgments in which each method's caption was chosen as the best description of the image.}
\label{tab:userstudy}
\end{table}

\section{Additional Qualitative Results}

\begin{figure}[t!]
\centering
\includegraphics[width=0.96\columnwidth]{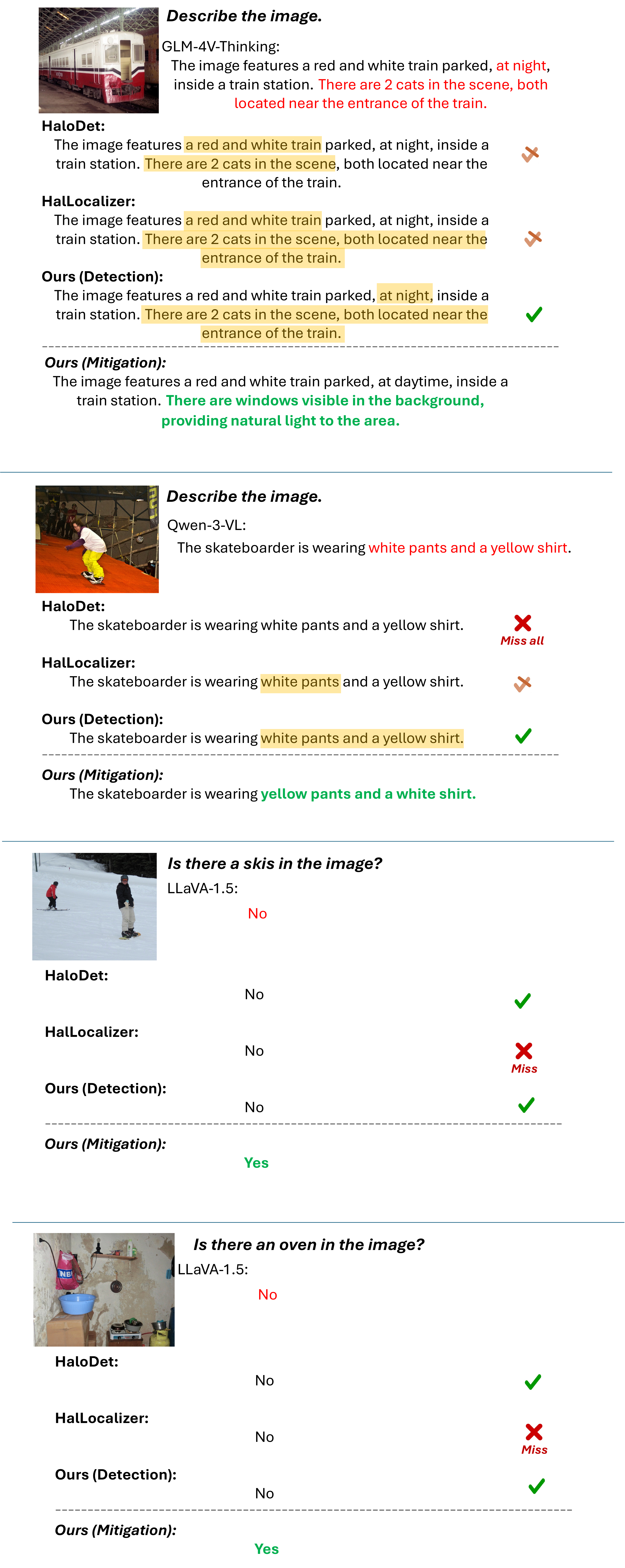}
\caption{\textbf{Additional detection and mitigation examples across backbones and hallucination types.} For each example we show the backbone response with the tokens flagged as hallucinated by HaloDet, HalLocalizer, and \ourmethod{} (Detection), followed by \ourmethod{}'s corrected generation (Mitigation). The cases cover object hallucination (an invented ``2 cats''), attribute hallucination (swapped colors and ``night''\,$\to$\,``daytime''), and object-presence errors on POPE questions (``skis'' and ``oven'' denied though present). \ourmethod{} localizes and repairs hallucinations that the baseline detectors miss, on GLM-4.1V, Qwen-3-VL, and LLaVA-1.5.}
\label{fig:qual_supp}
\end{figure}

\ourmethod{} can be applied across backbones and across hallucination types. Figure~\ref{fig:qual_supp} collects additional detection-and-mitigation examples on three backbones (GLM-4.1V, Qwen-VL, and LLaVA-1.5) that span the error modes we target: \emph{object} hallucination (a phantom ``2 cats'' invented in an otherwise empty train station), \emph{attribute} hallucination (swapped colors and lighting, such as ``white pants and a yellow shirt'' or ``at night''), and \emph{object-presence} errors on POPE-style questions, where the backbone denies an object (``skis'', ``oven'') that is in fact present. In each case \ourmethod{} localizes the hallucinated span while the trained detectors HaloDet and HalLocalizer miss it; conditioned on that signal, our mitigation rewrites the response into a grounded one, removing the cats and correcting ``night'' to ``daytime'', swapping the colors back, or answering ``Yes''. The presence cases are the most striking: the backbone verbally denies an object that its own internal state encodes as present, and \ourmethod{} recovers it directly from that trace.

\section{Additional Ablation Results}
We present additional ablation studies that complement the results reported in the main paper.

\begin{table}[t!]
\centering\small
\begin{tabular}{@{}lcc@{}}
\toprule
Attention edge features & F1\textsubscript{M} & F1\textsubscript{IoU} \\
\midrule
Mean over heads (\ourmethod{}) & \textbf{63.1} & \textbf{52.4} \\
All heads, single layer & 61.9 & 51.8 \\
All heads, all layers & 62.5 & 50.8 \\
\bottomrule
\end{tabular}
\caption{\textbf{Head averaging on MHALO.} Collapsing attention to the mean over heads (one scalar per edge) beats retaining all heads as per-edge features with a learned gate, which adds parameters and $L\!\times\!H$ features per edge without improving detection.}
\label{tab:attn_agg}
\end{table}

\subsection*{Budget}
\ourmethod{} keeps the trace graph compact by bounding two quantities: the number of \emph{nodes} it retains (at most $200$ image and $64$ query patches, ranked by the attention they receive from the response), and the number of \emph{edges} each response token draws from every modality. (1) Visual budget: Sweeping the image-node cap ($\{100,200,400,600\}$) and the per-token image edges ($k_{\text{img}}\!\in\!\{8,16,32,64\}$) leaves F1 essentially flat (F1\textsubscript{M} $62.1$--$63.0$, no trend): a small set of the most-attended patches already carries the visual evidence, and top-$16$ image edges per token suffice.
(2) Query/response edge budget: Varying the per-token edges to query tokens ($k_{\text{qry}}\!\in\!\{4,8,12,16\}$) and to preceding response tokens ($k_{\text{resp}}\!\in\!\{4,8,12\}$) leaves F1 within seed noise (F1\textsubscript{M} 62.4--63.2): a handful of edges to the most-attended context is enough; the defaults are top-$12$ query and top-$8$ response edges.

\subsection*{Head Averaging Ablation}
\ourmethod{} summarizes each attention edge by the \emph{mean} attention over heads. Table~\ref{tab:attn_agg} compares this against keeping all heads as separate per-edge features (with a learned gate over them), evaluated under an identical protocol. Averaging over heads is both the most accurate and the lightest: retaining every head adds an $L\!\times\!H$-dimensional feature per edge plus a gating layer, yet slightly \emph{lowers} detection, whether the heads are read at the single mid-network layer or across all layers.

\subsection*{Streaming Read-out} Making the response GRU unidirectional, as token-by-token deployment requires, costs ${\sim}2$ F1\textsubscript{M} on teacher-forced MHALO ($63.2\!\to\!61.2$, F1\textsubscript{IoU} tied), and nothing where it is actually used: on the model's own generations the streaming and bidirectional detectors tie (CHAIR detection F1 $32.6$ vs.\ $32.2$).

\end{document}